\documentclass[dvipsnames,format=sigconf]{acmart}

\usepackage{booktabs} 

\usepackage{algorithm}
\usepackage[noend]{algorithmic}
\usepackage[algo2e,ruled,vlined,linesnumbered]{algorithm2e}

\usepackage{caption}
\usepackage{subcaption}
\usepackage{multirow}
\usepackage{adjustbox}
\usepackage{tabularx}

\setcopyright{rightsretained}

\begin{document}
\title{Automated design of relocation rules for minimising energy consumption in the container relocation problem}

\author{Marko Đurasević}
 \affiliation{%
   \institution{University of Zagreb, Faculty of Electrical Engineering and Computing}
 \city{Zagreb} 
 \country{Croatia} 
 }
\email{marko.durasevic@fer.hr}

\author{Mateja Đumić}
 \affiliation{%
   \institution{Department of Mathematics, Josip Juraj Strossmayer University of Osijek}
 \city{Osijek} 
 \country{Croatia} 
 }
\email{mdjumic@mathos.hr}

\author{Rebeka Čorić}
 \affiliation{%
   \institution{Department of Mathematics, Josip Juraj Strossmayer University of Osijek}
 \city{Osijek} 
 \country{Croatia} 
 }
\email{rcoric@mathos.hr}

\author{Francisco J. Gil-Gala}
 \affiliation{%
   \institution{University of Oviedo. Department of Computing}
   \city{Gijón}
   \country{Spain}}
\email{giljavier@uniovi.es}

\renewcommand{\shorttitle}{}

\begin{abstract}
The container relocation problem is a combinatorial optimisation problem aimed at finding a sequence of container relocations to retrieve all containers in a predetermined order by minimising a given objective. 
Relocation rules (RRs), which consist of a priority function and relocation scheme, are heuristics commonly used for solving the mentioned problem due to their flexibility and efficiency. 
Recently, in many real-world problems it is becoming increasingly important to consider energy consumption. 
However, for this variant no RRs exist and would need to be designed manually. 
One possibility to circumvent this issue is by applying hyperheuristics to automatically design new RRs. 
In this study we use genetic programming to obtain priority functions used in RRs whose goal is to minimise energy consumption. 
We compare the proposed approach with a genetic algorithm from the literature used to design the priority function.
The results obtained demonstrate that the RRs designed by genetic programming achieve the best performance. 
\end{abstract}

%
%




\maketitle

\sloppy
\section{Introduction}
The container relocation problem (CRP), is a combinatorial optimisation problem with applications in warehouse and yard management \cite{Jovanovic2019}. 
Due to the limited space, containers are usually stacked one atop another and/or side by side. 
This way blocks are formed that have stacks (width), a number of tiers (height), and a number of bays (length).
The objective is to retrieve and load all containers from the yard in a predetermined order.
However, a container can be retrieved only if it is located on the top of its stack. 
If there are containers on top of the one that needs to be retrieved, they first need to be relocated to other stacks. 

During the years, many heuristics and metaheuristics were proposed to solve this problem \cite{Kim2006, Wu2010, Zhu2012, Cifuentes2020}. 
These methods are computationally expensive and require a substantial amount of time to obtain solutions for larger problem sizes.
Therefore, simple heuristic methods, called relocation rules (RRs), are proposed in the literature to solve CRP \cite{Wu2010, Caserta2012}.
RRs construct the solution incrementally by determining which relocation should be performed based on the current system information.
For that purpose, RRs use a priority function (PF) to rank all possible relocations and select the best one. 
Since manually designing such PFs is difficult, certain studies investigated the possibility of automatically designing them \cite{Hussein2012, Durasevic2022}. 

Due to the growing environmental concerns that arise today, optimising energy related criteria is becoming increasingly important in various optimisation problems, such as vehicle routing \cite{Erdelic2019} or various scheduling problems \cite{ Durasevic2022a}. 
However, in CRP the energy consumption criterion did not receive much attention, with only a few studies focusing on optimising it either directly \cite{Hussein2012}, or as a part of the total cost objective \cite{LOPEZPLATA2019436}.
Thus, there is a lack of RRs that could be used to efficiently optimise this criterion. 

To close this gap, we examine the application of hyperheuristics to generate RRs appropriate for optimising the total energy consumption during the retrieval process of containers. 
Although this problem was tackled in \cite{Hussein2012}, the authors manually defined the mathematical expression of the priority function used to rank all the relocations, and used a GA to optimise certain parameters in that expression.
As such, the approach is limited in a sense that the structure of the priority function still needs to be defined manually.
Therefore, we revisit this problem and apply GP as a hyperheuristic to generate PFs of an arbitrary structure. 
We use the same information as the authors in \cite{Hussein2012} to make a fair comparison between the methods, and show that PFs designed by GP construct significantly better solutions than the previously proposed GA. 
The contributions of this paper can be outlined as follows:
\begin{enumerate}
    \item develop a GP based hyperheuristic method to optimise the total energy consumption for CRP;
    \item compare priority functions for relocation rules evolved by GA and GP.
\end{enumerate}


\section{CRP problem description}
We consider the single bay CRP in which the bay consists of $S$ stacks with $H$ tiers. 
Every stack has a height denoted with $h(S)$ that has to be less or equal to the maximum height $H$. 
There are $C$ containers in the bay and a single crane that can move one container at a time.
Each container has a different priority, which denotes the order of their retrieval from the yard. 
To solve the problem, two types of operations can be performed by the gantry crane, relocation and retrieval. 
Relocation moves a container from the top of one stack to another, which can be done only if the stack to which the relocation is being made has a height smaller than $H$. 
The second operation, retrieval, picks a container from the top of the stack and moves it to the truck used for loading, which is located at position 0 denoting the beginning of the bay. 


Each container in the bay has an ID that determines in which order they need to be retrieved. 
The container with the smallest ID in the yard is the one that needs to be retrieved next, and is called the \emph{target container}. 
If the target container is not located at the top of its stack, it is required to relocate all the containers above it to different stacks.
The stack from which the container is moved is called the \emph{origin stack}, whereas the stack to which the container is moved is called the \emph{destination stack}.
All relocation and retrieval sequences that guarantee the crane can retrieve every container in a predetermined order denote feasible CRP solutions. 
The goal is to find a sequence that minimises a given objective. 
In this study we optimise the total energy consumed while retrieving all containers from the yard, which can be defined as \cite{Hussein2012}:
$$TEC=\sum_{m=1}^{M}W_m(h*h_m+l*l_m+x*x_m),$$
where:
\begin{itemize}
    \item $h$  ($l$)   -- energy consumed per ton for one tier hoisted (lowered)
    \item $x$ -- energy consumed per ton when moving the crane stack
    \item $h_m$ ($l_m$) -- tiers hoisted (lowered) during move $m$
    \item $x_m$ -- stacks crossed during move $m$
    \item $W_m$ -- moving weight of move $m$; $W_m=W_s+W_c$, where $W_s$ denotes the weight of the crane, and $W_c$ denotes the weight of the container moves (equals to 0 if crane was empty)
    \item $M$ -- number of moves required to retrieve all containers.
\end{itemize}
Based on \cite{Hussein2012}, values for $h$, $l$, $x$ are set to $0.9$, $0.02$ and $0.08$ respectively, while crane weight $W_s$ is equal to $0.5$ tons.

\section{Methodology}\label{section:methodology}
\subsection{Relocation rules}
Relocation rules (RRs) represent simple constructive heuristics that iteratively build the solution to CRP.
They consist of two parts - the relocation scheme (RS) and the priority function (PF) \cite{Durasevic2022}. 
RS takes care of problem constraints and creates a plan for container retrieval and relocation. 
If the container that needs to be retrieved next is on top of its stack, it is retrieved, otherwise, the containers above it must be moved to another stack to allow retrieval. 
RS uses PF to decide which stacks the containers above the target container should be moved to in order to relieve the target containers. 
This is done iteratively, one container at a time. 
RS determines the container that needs to be moved next, and PF assigns a numeric value to each stack to which the given container can be moved. 
Depending on PF, the container is moved to the stack for which the best value was determined.

RSs are simple algorithms that are defined manually. 
Based on the moves that are allowed we distinguish between the restricted and the unrestricted RS. 
In the restricted version, only containers located above the target container may be moved, while in the unrestricted version, there is no such restriction, i.e., all containers located on top of their stack may be moved. 

\subsection{Using GA and GP for developing PFs}

Designing a good PF manually is a challenging task, because of which several attempts to automate this process were performed. 
Partial automation was done in \cite{Hussein2012}, in which the authors manually defined a general expression with a certain number of free parameters that were optimised with GA. In a more recent work the entire PF was developed using GP \cite{Durasevic2022}, which achieved significantly better performance than several existing manually designed PFs.

In this work, we consider both ways to design PFs to optimise the total energy consumption for CRP. 
Both GA and GP use the same evolutionary scheme, with the the main difference being the representation of individuals. 
The GA uses a list of floating point numbers denoting the parameters it optimises, while GP uses the standard expression tree representation.
The evaluation is done using a fitness function that evaluates each individual on a set of problems and assigns a numerical value (fitness) to the individual. In each iteration, a 3-tournament selection is used, the two better of the selected individuals are used for crossover, and the worst is replaced by a newly created individual to which the mutation operator is applied with a certain probability. This is repeated until the maximum number of fitness function evaluations is reached.


In \cite{Hussein2012}, the authors propose the global retrieval heuristic (GRH) for restricted CRP with container weights to optimise total energy consumption.
In this study, we adapt GRH to also work with the unrestricted RS to test whether this can improve the results.
When deciding where to move the container, GRH uses a penalty function and selects the stack that received the lowest value.
The penalty function is given by expression (\ref{eq:penalty}), and the description of the variables can be found in the Tables \ref{tab:GRHSettings} and \ref{tab:algVariables}.
Table \ref{tab:algVariables} outlines the variables that the algorithm uses as inputs, while Table \ref{tab:GRHSettings} contains the free parameters that must be set before solving the problem and whose values are between 0 and 1.
The idea presented in the paper \cite{Hussein2012} is to apply a GA to determine the free parameters from Table \ref{tab:GRHSettings}.
The GA uses a simple floating-point encoding, where each individual consists of 12 real numbers, each denoting one of the parameters.


GP as a hyperheuristic achieves good results in automatic development of scheduling rules \cite{Branke2016, Nguyen2017} and has been successfully applied to the basic CRP problem \cite{Durasevic2022}, in which the total number of relocations and crane operation time were optimised. 
Encouraged by this, in this paper we apply GP to generate PFs to minimise the total energy consumption in CRP. 
To analyse how GP compares to the GA approach of \cite{Hussein2012}, GP uses the same system information to construct the PF. 
This means that the terminal set of GP comprises of the variables given in Table \ref{tab:algVariables} (except $A_1, A_3$ and $A_4$). 
The set of functions used in the development of the penalty function consists of addition, subtraction, multiplication and protected division (returns 1 if the divisor is close to 0).

\begin{table}[htbp]
  \centering
  \caption{GRH settings \cite{Hussein2012}}
    \begin{tabularx}{0.45\textwidth}{lX}
    \toprule
    Parameter & Description \\
    \midrule
    $\alpha$ / $\beta$ / $\gamma$   & importance of minimising hoisting/lowering/trolleying \\
    $P_1$  & importance of minimising hoisting, lowering, and trolleying of heavy (versus light) containers  \\
    $\delta$ / $\epsilon$   &importance of minimising/delaying rehandling \\
    $P_2$ / $P_3$    & importance of minimising/delaying rehandling of heavy (versus light) containers \\
    $\eta$  & importance of tightness \\
    $\theta$  & importance of moving containers closer to the truck lane \\
    $P_4$  & importance of moving heavy (versus light) containers closer to truck lane \\
    $\mu$  & importance of keeping stack heights low \\
    \bottomrule
    \end{tabularx}
  \label{tab:GRHSettings}
\end{table}

\begin{table}[htbp]
  \centering
  \caption{Variables contained in the penalty score function whose values are calculated when the container $c$ with weight $W_c$ is reshuffled to the destination stack $s$ \cite{Hussein2012}}
    \begin{tabularx}{0.45\textwidth}{lX}
    \toprule
    Variable & Description \\
    \midrule
    $h_s$ / $l_s$ / $x_s$   & number of tiers hoisted/lowered/trolleyed when moving reshuffled container to stack $s$ \\
    $r_s$  &  binary variable that equals 1 if the container c must be reshuffled again if it is placed on the stack s, otherwise it equals 0.  \\
    $t_s$  & lowest numbered container in stack $s$ \\
    $g_s$  & tightness, calculated with formula $g_s=\frac{t_s-c-1}{C}$ \\
    $k_s$  & amount of trolley movement away from truck lane if container c reshuffled to stack $s$; is equal to $0$ if trolley moves toward truck lane when container reshuffled to stack $s$ \\
    $n_s$  & number of containers in stack $s$ \\
    \midrule
    $A_1$ & $1+\frac{W_c}{W_{max}}\cdot 10\cdot P_1$ \\
    $A_3$ & $A_3 = 1+\frac{W_c}{W_{max}}\cdot 10\cdot P_3$\\
    $A_4$ & $1+\frac{W_c}{W_{max}}\cdot 10\cdot P_4$\\
    \bottomrule
    \end{tabularx}
  \label{tab:algVariables}
\end{table}

\begin{equation}\label{eq:penalty}
\begin{aligned}
 Penalty_s ={} &\alpha\left(\frac{h_s}{mxHeight}\right)^{A_1} + \beta\left(\frac{l_s}{mxHeight}\right)^{A_1} +\gamma\left(\frac{x_s}{S}\right)^{A_1} \\
 & + \delta r_s \left(\frac{W_c}{W_{max}}\right)^{10P_1} + \epsilon r_s \left(\frac{c-t_s}{C}\right)^{A_3} + \eta(1-r_s)g_s \\
 & + \theta \left(\frac{k_s}{S}\right)^{A_4} + \mu \left(\frac{n_s}{mxHeight}\right)   
\end{aligned}
\end{equation}


\section{Experimental setup} \label{section:experiments}

To test the performance of GRH and GP evolved PFs for RRs, the Caserta~\cite{Caserta2011} and Zhu~\cite{Zhu2012} datasets are used. 
These original instances are used as the test set, whereas additional instances were generated to be used for training GP and GRH. In order to be able to use the original instances from these two sets with energy criteria, an additional weight with an uniform distribution from 1 to 30 was generated for each container, as was done in \cite{Hussein2012}.
The adapted problem instances can be obtained from \url{http://www.zemris.fer.hr/~idurasevic/CRP/CRP.7z}. 

Both GP and GA use a population of 1\,000 individuals, mutation probability of 0.3 for the restricted and 0.1 for the unrestricted RRs, and 50\,000 function evaluations. 
The maximum tree depth was set to 5 in GP.
GP used the subtree, uniform, context preserving, size fair, and one point crossover operators, as well as the subtree, hoist, node complement, node replacement, permutation, and shrink mutation operators \cite{Poli2008}. 
The GA used several well known genetic operators, like arithmetic, SBX, BLX-$\alpha$, and others.
For mutation, the uniform mutation operator is used, which generates a random number from the interval $[0,1]$.
In cases when several crossover or mutation operators are defined, a random one is selected and applied for each time the operator needs to be invoked. 

To obtain a notion on the performance of the algorithms, GP and GA were executed 30 times to evolve RRs using the training set. 
The best RR obtained in each execution is evaluated on the test set and the total consumed energy for each of these 30 rules is determined. 
To test whether the obtained results are statistically significant, the Kruskal-Wallis test with the Dunn post hoc test and Bonferroni correction method was used. 
The obtained differences are considered significant if a p-value below 0.05 was obtained.


\section{Results}\label{section:results}

\begin{figure}[hbt!]
\centering
\begin{subfigure}[b]{0.7\linewidth}
    \centering
    \includegraphics[width=0.99\linewidth]{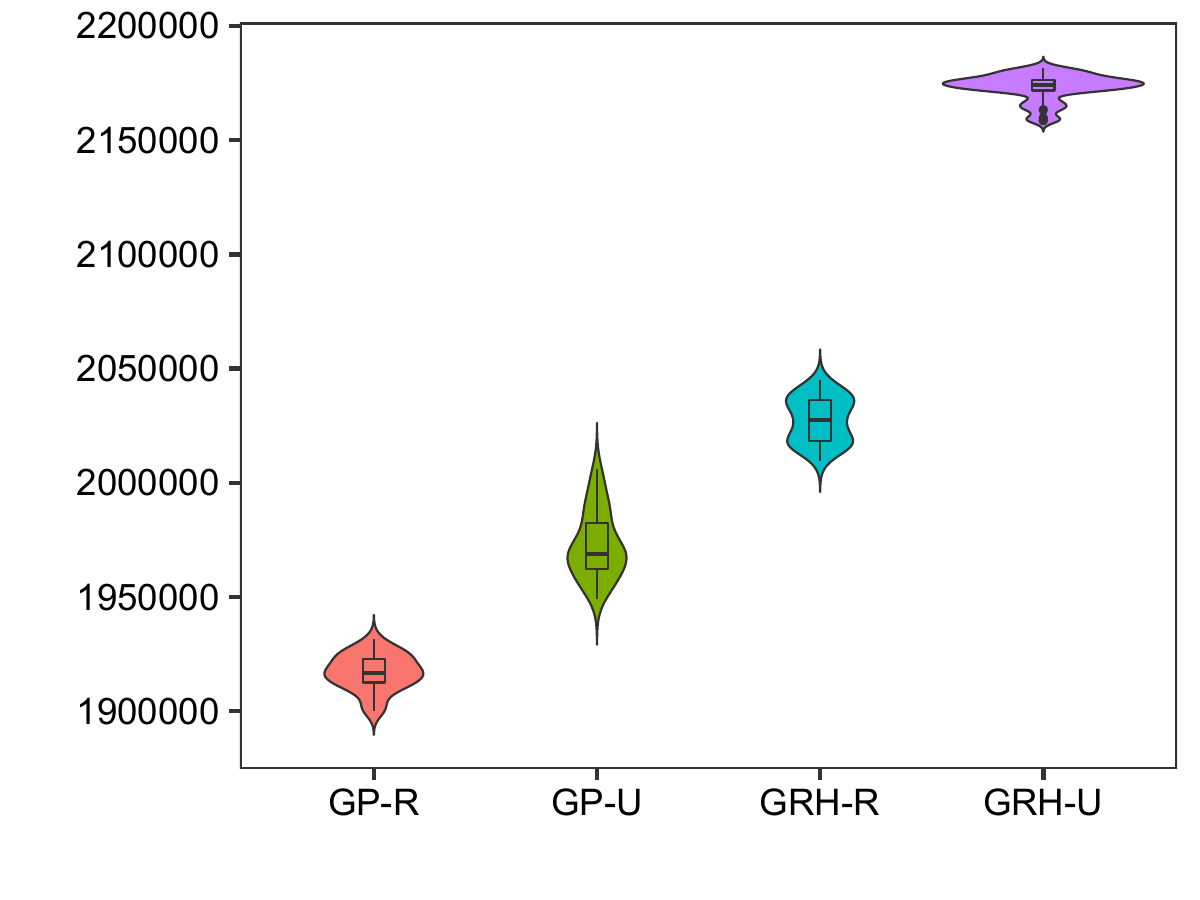}
    \caption{Caserta dataset}
\end{subfigure}
\begin{subfigure}[b]{0.7\linewidth}
    \centering
    \includegraphics[width=0.99\linewidth]{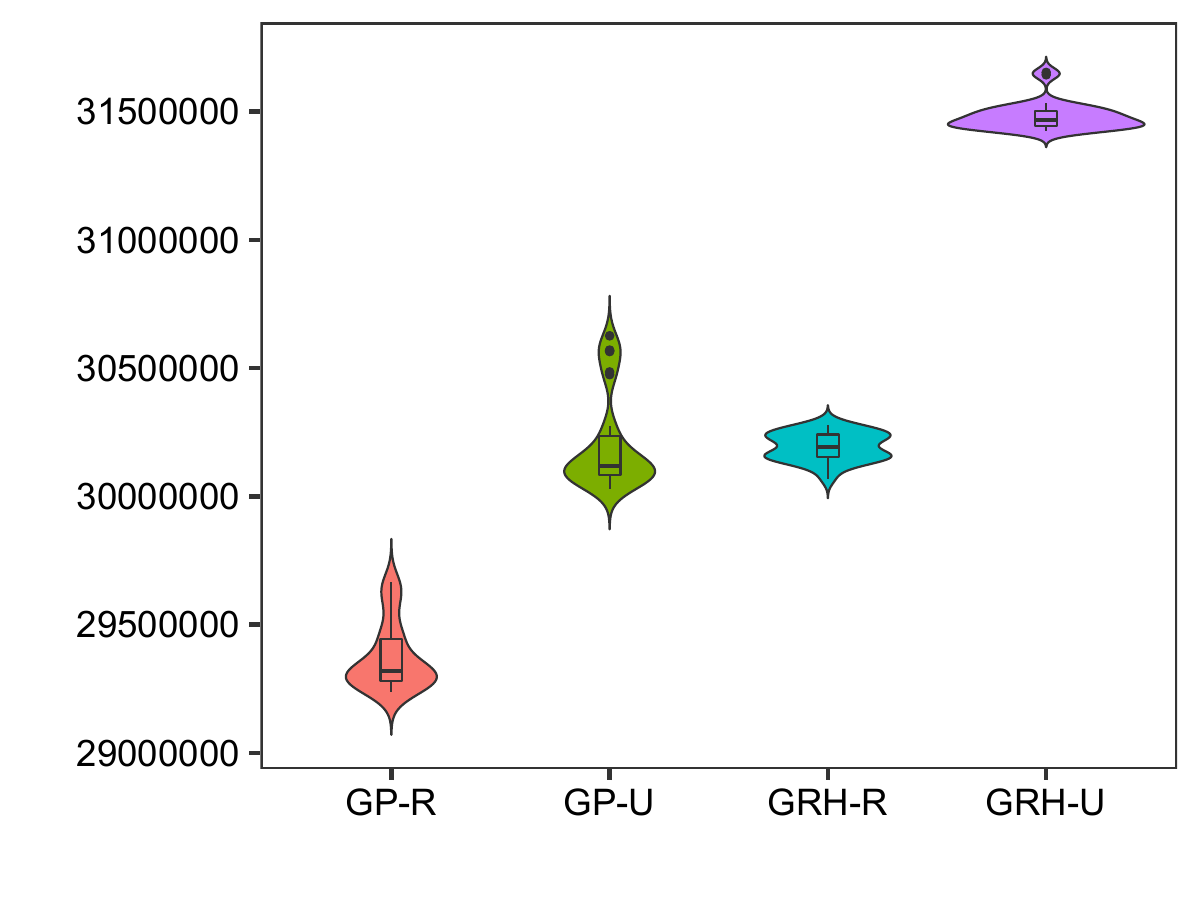}
    \caption{Zhu dataset}
\end{subfigure}
\vspace{-0.2cm}
\caption{\label{fig:boxplots} Results of automatically designed RRs.}
\end{figure}

Figure \ref{fig:boxplots} outlines the results obtained for RRs generated by GP and GRH.
By comparing the two methods used to automatically design RRs, we see that GP evolved PFs consistently achieve a better minimum and median values of the results on both datasets. 
The first thing to notice is that the restricted versions (marked in the figure with -R next to the name of the approach) of the RRs consistently perform better than their unrestricted variants (marked in the figure with -U next to the name of the approach).
The reason why this happens is that the unrestricted version introduces additional moves that are performed, which ultimately increases the total consumed energy. 
As we see, this increase is quite substantial and therefore leads to significant deterioration of the results. 
Therefore, we can conclude that for this optimisation criteria the unrestricted version of RRs is not appropriate. 

If we compare the rules obtained with GP and the GRH, we see that the rules generated by GP perform better in almost all cases.
This is most evident in the restricted variant, where even the worst solution obtained with GP outperforms the best solution obtained with GRH.
The only discernible advantage of GRH is that the results are less dispersed than with GP. However, as these results are generally worse, this has no obvious advantage.
On average, the results obtained with GP are about 5\% better than those obtained with GRH. The statistical tests showed that restricted RRs evolved by GP perform significantly better than all other RR variants, thus confirming its superiority. 
Furthermore, the restricted RR variants always perform significantly better than their unrestricted counterparts, proving that the restricted variants are preferable for optimising this criterion.
Finally, GP evolved RRs always perform better than GA evolved rules, except in one case in the Zhu dataset, where only GP-U and GRH-R perform equally well.
Based on these results, we can conclude that GP is more appropriate for designing new RRs, compared to GA coupled with GRH.

\section{Conclusions and Future work}\label{section:conclusion}

In this paper, the application of GP to automatically generate RRs for solving CRP with the aim of minimising total energy consumption was investigated.
The method was tested on an extensive set of experiments and compared with GRH, where the parameters for RR are adjusted using a GA.
The experimental results show that RRs designed using GP significantly outperform those designed using GRH.
This shows the versatility of GP in obtaining high quality RRs for CRPs with non-standard criteria, especially considering that it uses the same system properties as GRH.
Thus, we see that the additional flexibility of GP to freely design the expression of PF gives it the ability to obtain better solutions than rules where the structure is manually defined and only the corresponding coefficients are optimised.
Based on these results, we conclude that GP can be used to generate effective RRs for new problem variants of CRP.

In the future work we will propose several new terminal nodes for minimising total energy consumption.
Furthermore, it is planned to optimise the energy consumption in a multi-objective scenario with other criteria like the total number of relocations. 
Finally, the model considering energy minimisation will be extended to other CRP variants including multiple bays and duplicate containers. 
    
 \begin{acks}
 This research has been supported by the Croatian Science Foundation under project IP-2019-04-4333 and the Spanish State Agency for Research (AEI) under research project PID2019-106263RB-I00.
\end{acks}

\bibliographystyle{ACM-Reference-Format}
\bibliography{mybibfile} 

\end{document}